\documentclass[10pt,twocolumn,letterpaper]{article}

\usepackage{cvpr}
\usepackage{times}
\usepackage{epsfig}
\usepackage{graphicx}
\usepackage{amsmath}
\usepackage{amssymb}
\usepackage{multirow}

\graphicspath{{pdf/}{jpeg/}}
\DeclareGraphicsExtensions{.pdf,.jpeg,.png}
\usepackage[caption=false,labelfont=sf]{subfig}

\cvprfinalcopy

\usepackage[breaklinks=true,bookmarks=false]{hyperref}

\def\cvprPaperID{1216} 

\begin{document}



\title{Unsupervised Video Object Segmentation with Distractor-Aware Online Adaptation}




\author{Ye Wang$^{1}$, Jongmoo Choi$^{1}$, Yueru Chen$^{1}$, Siyang Li$^{2}$, Qin Huang$^{3}$, Kaitai Zhang$^{1}$, \\
Ming-Sui Lee$^{4}$, and C.-C. Jay Kuo$^{1}$\\
$^{1}$University of Southern California, $^{2}$Google AI Perception,
$^{3}$Facebook, $^{4}$National Taiwan University \\
{\tt\small \{wang316,jongmooc,yueruche\}@usc.edu, siyang@google.com, huginhuang@fb.com,}\\ {\tt\small kaitaizh@usc.edu, mslee@csie.ntu.edu.tw, cckuo@sipi.usc.edu}
}


\maketitle

\begin{abstract}

Unsupervised video object segmentation is a crucial application in video analysis without knowing any prior information about the objects. It becomes tremendously challenging when multiple objects occur and interact in a given video clip. In this paper, a novel unsupervised video object segmentation approach via distractor-aware online adaptation (DOA) is proposed. DOA models spatial-temporal consistency in video sequences by capturing background dependencies from adjacent frames. Instance proposals are generated by the instance segmentation network for each frame and then selected by motion information as hard negatives if they exist and positives. To adopt high-quality hard negatives, the block matching algorithm is then applied to preceding frames to track the associated hard negatives. General negatives are also introduced in case that there are no hard negatives in the sequence and experiments demonstrate both kinds of negatives (distractors) are complementary. Finally, we conduct DOA using the positive, negative, and hard negative masks to update the foreground/background segmentation. The proposed approach achieves state-of-the-art results on two benchmark datasets, DAVIS 2016 and FBMS-59 datasets. 

   
\end{abstract}

\section{Introduction}

Video object segmentation (VOS) aims to segment foreground objects from complex background scenes in video sequences. There are two main categories in existing VOS methods: semi-supervised and unsupervised. Semi-supervised VOS algorithms \cite{marki2016bilateral,tsai2016video,voigtlaender2017online,caelles2017one,perazzi2017learning,hu2018videomatch} require manually annotated object regions in the first frame and then automatically segment the specified object in the remaining frames throughout the video sequence. Unsupervised VOS algorithms \cite{koh2017primary,tokmakov2017learning,jain2017fusionseg,li2018instance,li2018unsupervised,hu2018unsupervised} segment the most conspicuous and eye-attracting objects without prior knowledge of these objects in the video. We focus on the unsupervised setting in this paper.

Motion information is a key factor in unsupervised video object segmentation since it attracts people's and animals' attention. \cite{lee2011key} initializes the segments from the motion saliency and then propagates to the remaining frames. However, the initialized motion saliency regions are not accurate enough and it could fail when multiple adjacent objects are moving in the scene. More recently, deep learning models have also been applied to automatically segment the moving objects with motion cues. FSEG \cite{jain2017fusionseg} trains a dual-branch fully convolutional neural network, which consists of an appearance network and a motion network, to jointly learn the object segmentation and optical flow. Direct fusion of the optical flow and object segmentation cannot be able to build the correspondence between foreground and motion patterns.

\begin{figure}[t]
\captionsetup[subfigure]{labelformat=empty}
\centering 
\subfloat{\includegraphics[width=0.24\linewidth]{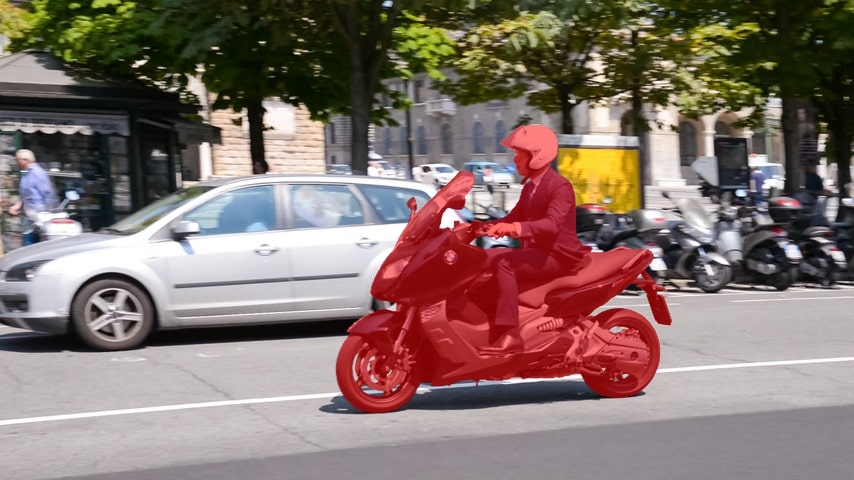}} \hfil 
\subfloat{\includegraphics[width=0.24\linewidth]{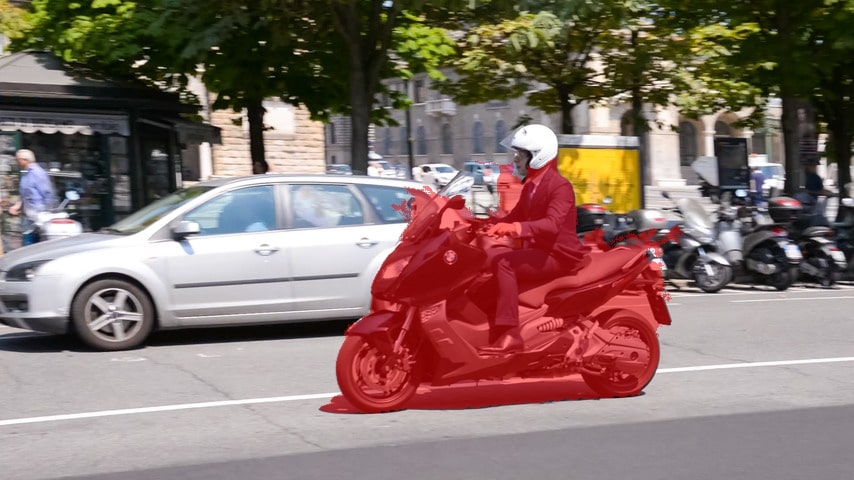}} \hfil 
\subfloat{\includegraphics[width=0.24\linewidth]{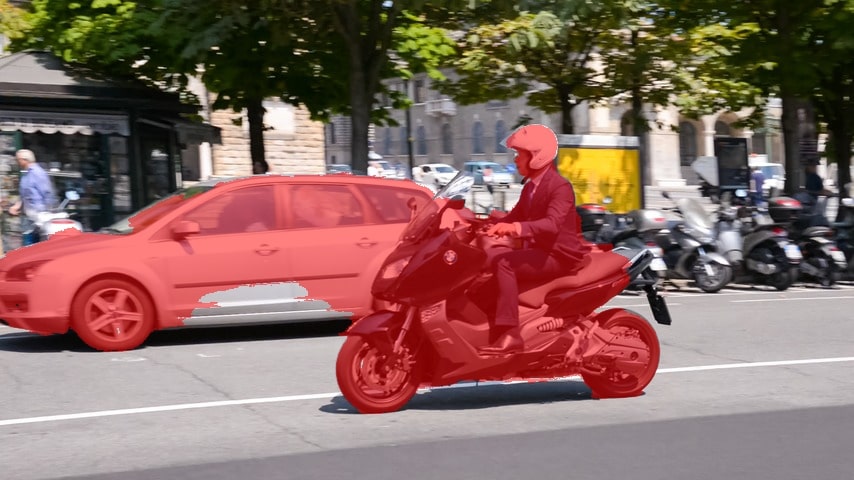}} \hfil 
\subfloat{\includegraphics[width=0.24\linewidth]{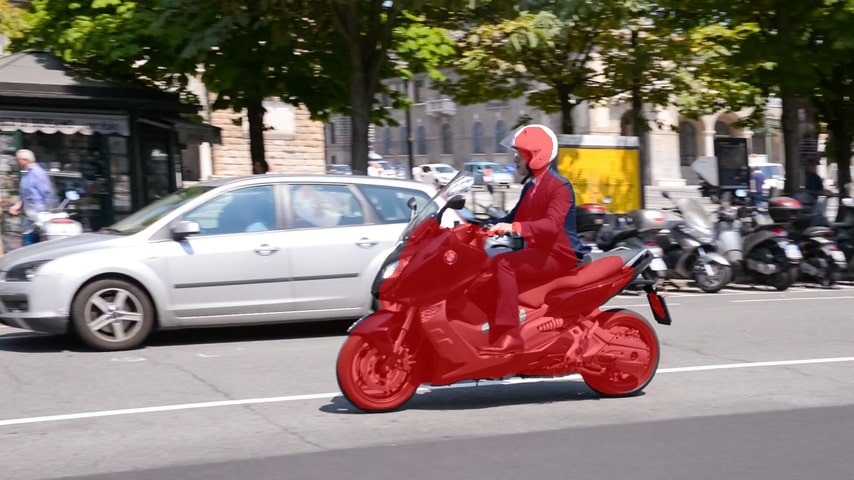}} \hfil  \\
\vspace{-0.12in}
\subfloat{\includegraphics[width=0.24\linewidth]{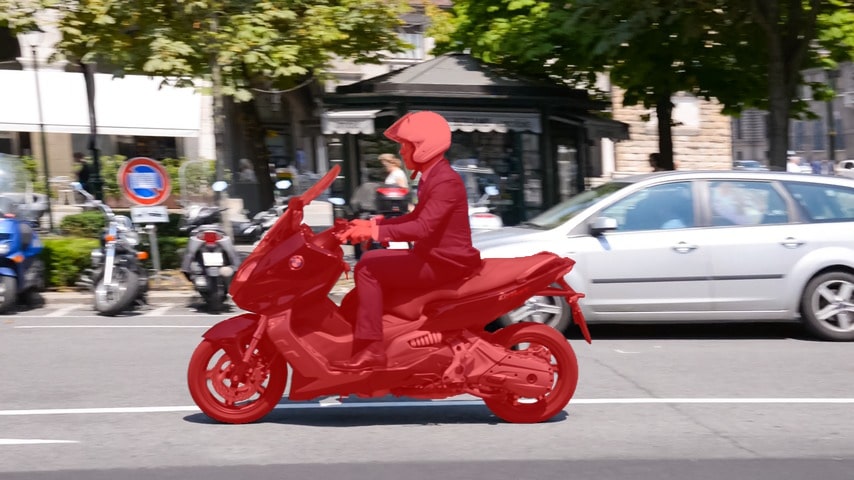}} \hfil 
\subfloat{\includegraphics[width=0.24\linewidth]{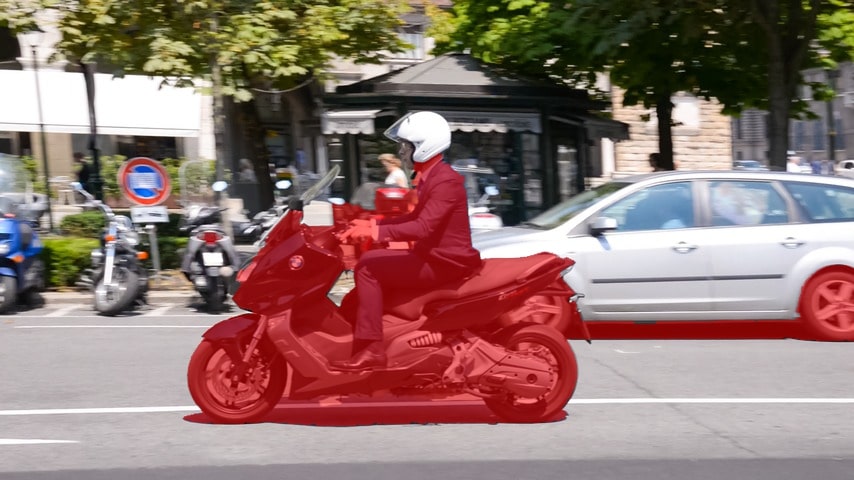}} \hfil 
\subfloat{\includegraphics[width=0.24\linewidth]{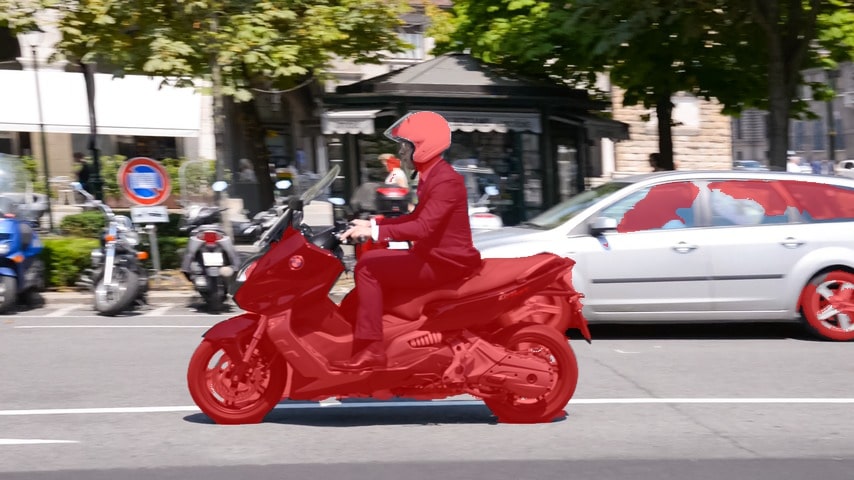}} \hfil 
\subfloat{\includegraphics[width=0.24\linewidth]{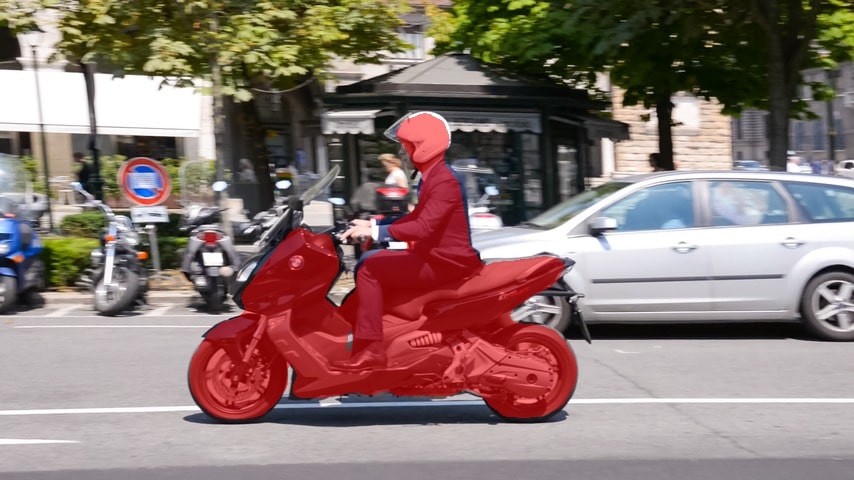}} \hfil  \\
\vspace{-0.12in}
\subfloat{\includegraphics[width=0.24\linewidth]{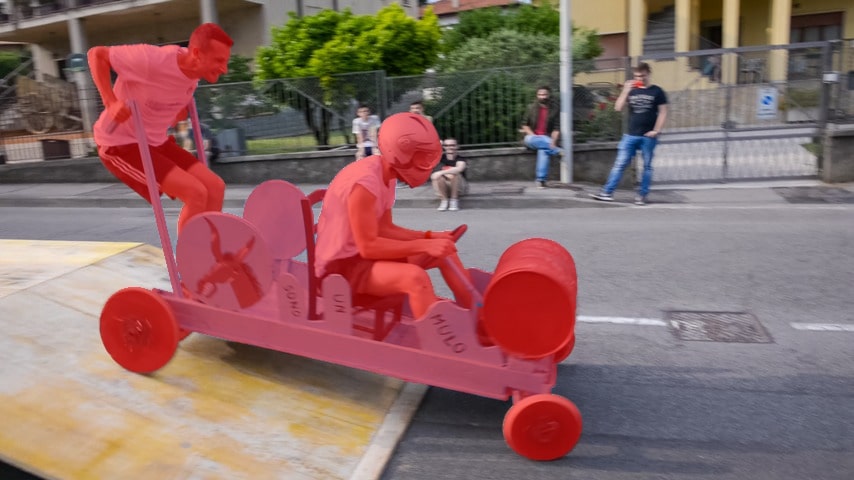}} \hfil 
\subfloat{\includegraphics[width=0.24\linewidth]{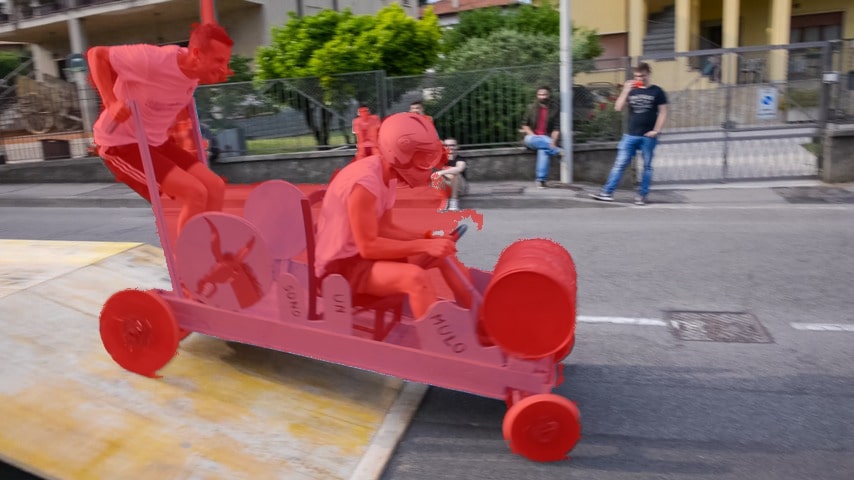}} \hfil 
\subfloat{\includegraphics[width=0.24\linewidth]{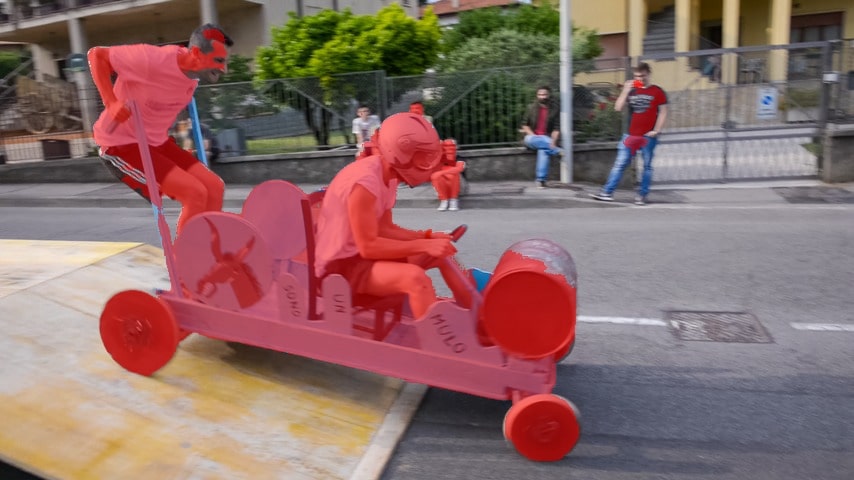}} \hfil 
\subfloat{\includegraphics[width=0.24\linewidth]{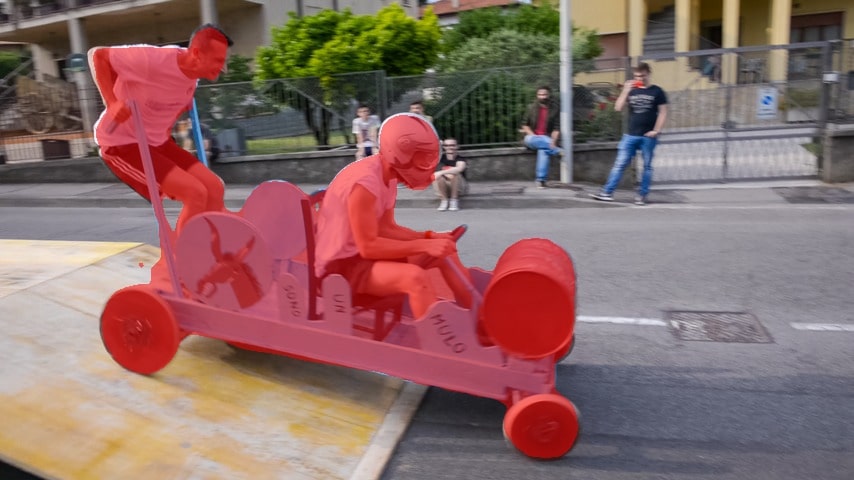}} \hfil  \\
\vspace{-0.12in}
\subfloat[Ground truth]{\includegraphics[width=0.24\linewidth]{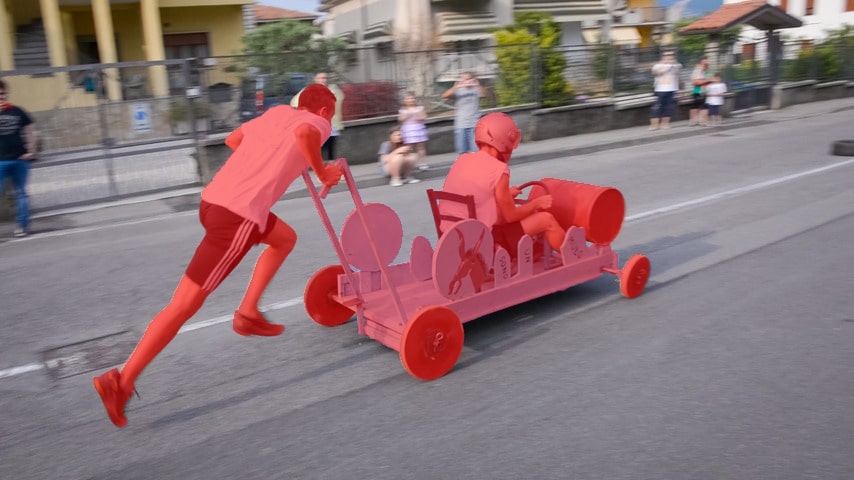}} \hfil 
\subfloat[ARP]{\includegraphics[width=0.24\linewidth]{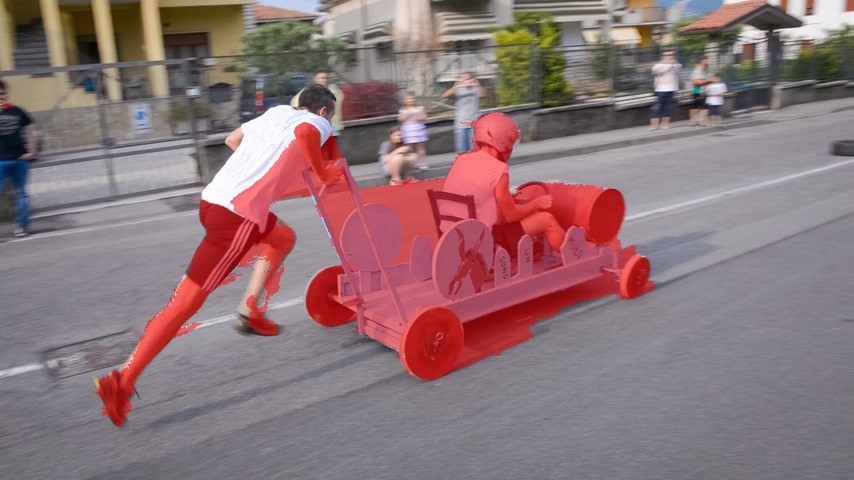}} \hfil 
\subfloat[PDB]{\includegraphics[width=0.24\linewidth]{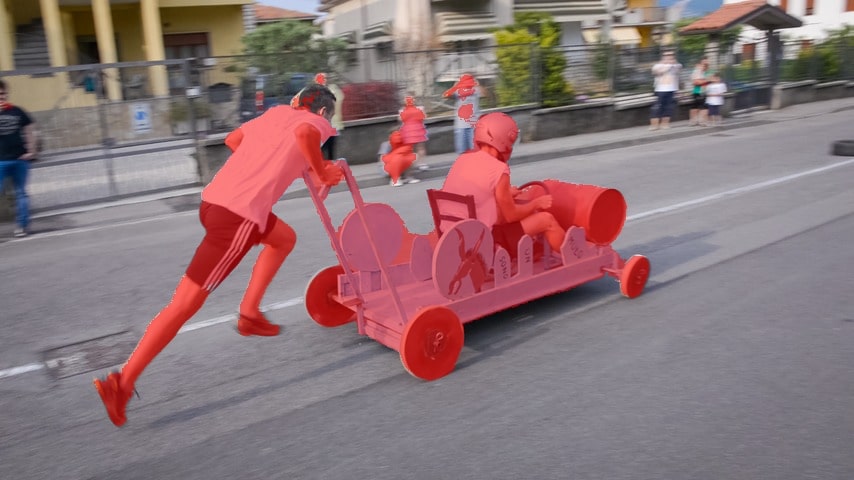}} \hfil 
\subfloat[Ours]{\includegraphics[width=0.24\linewidth]{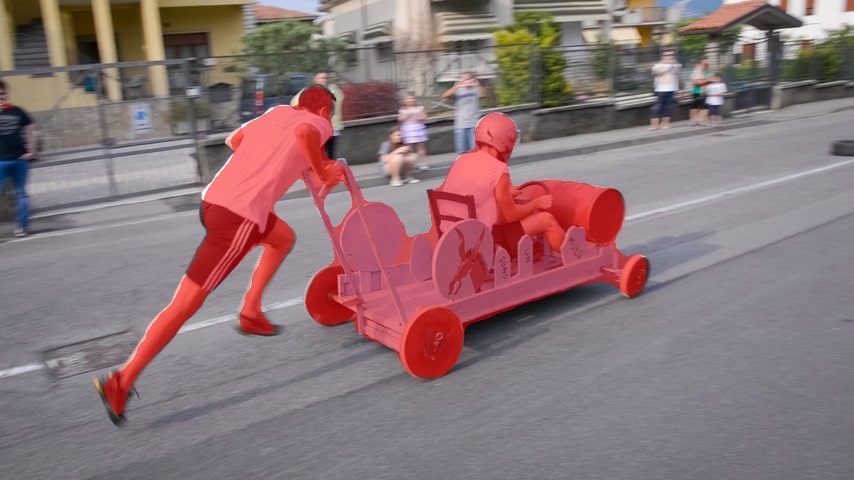}} \hfil  \\
{}
\caption{Example results of ground truth, ARP \cite{koh2017primary}, PDB \cite{song2018pyramid} and the proposed method. Distractors in the background lead to incorrect segmentations (second and third column) due to the similar features between the foreground and background objects. We exploit distractor-aware online adaptation approach to learn from the hard negatives to avoid mis-classifying background objects as foreground. Best viewed in color with 4$\times$ zoom.}\label{fig:example}
\end{figure}

In this paper, we aim at generating accurate initialized foreground mask by leveraging both optical flow and instance segmentation, which is further refined by performing distractor-aware online adaptation to generate consistent foreground object regions. Specifically, the instance segmentation algorithm is first applied to roughly segment the objectiveness masks, and then the foreground object of the first frame is further grouped and selected by leveraging optical flow. We call this accurate foreground mask ``pseudo ground truth''. Finally, one-shot VOS \cite{caelles2017one} could be applied to propagate the prediction to the remaining frames of the given video. However, the main problems for one-shot VOS are the unstable boundaries and treating static and moving objects the same. To this end, distractor-aware online adaptation is proposed to utilize the motion information through the whole video to improve the inter-frame similarity of the consecutive masks and avoid the motion error propagation by exploiting the fusion of motion and appearance.

Inspired by DOA, we erode the prediction mask from the previous frame and mark it as positive examples for the current frame. As it is known to all, consistently paying attention to a previously seen background object, when it meets with the target foreground object, segmentations should be easily distinguishable. The instance proposals that are not covered by motion masks in the first frame are treated as hard negatives. We apply block matching algorithm to find corresponding blocks from adjacent frames. If the intersection-over-union of the block and instance proposals is larger than a certain threshold, the instance proposals are considered as hard negative examples. Therefore hard negative attention-based adaptation is applied to update the network. Sample results are illustrated in Figure \ref{fig:example}.

\begin{figure*}[t]
\captionsetup[subfigure]{}
\centering 
\subfloat{\includegraphics[width=0.9\linewidth]{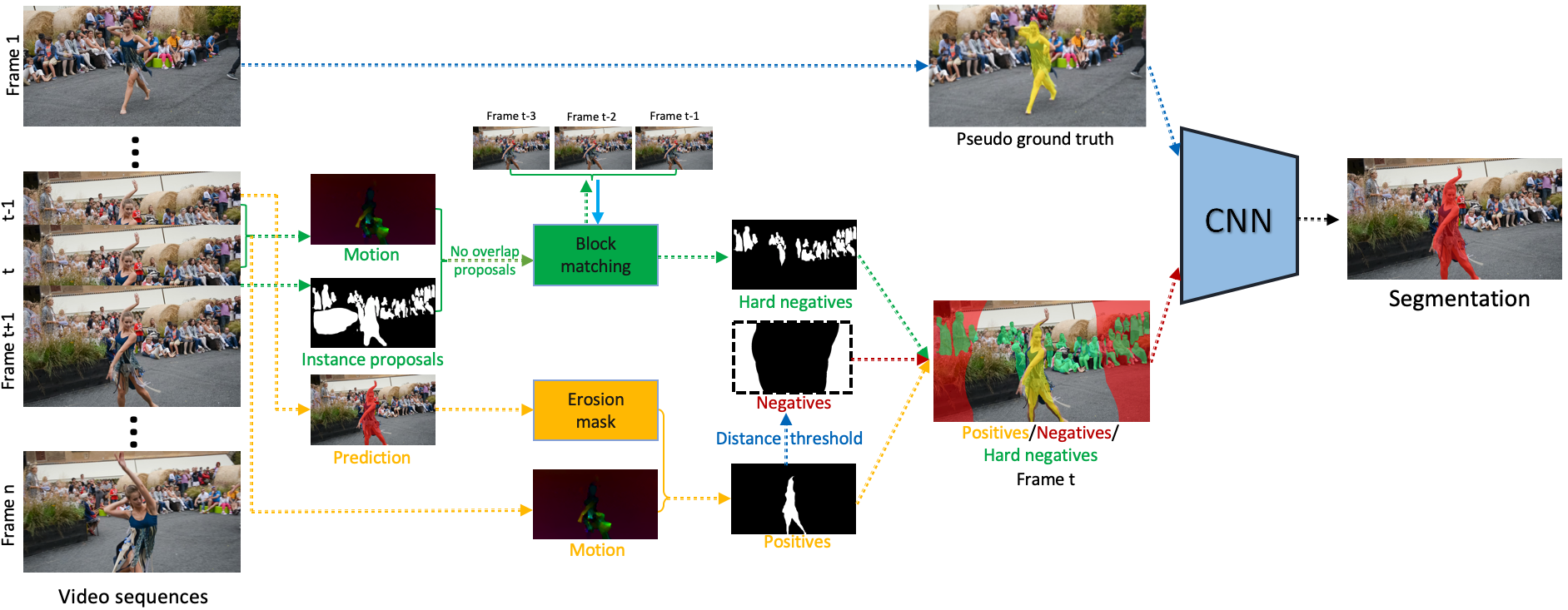}} \hfil
\caption{Overview of the proposed method. Instead of directly applying static image segmentation to video object segmentation, an online adaptation approach is proposed by detecting distractors (negatives and hard negatives). Both the appearance and motion cues are utlized to generate positives, negatives, and hard negatives. Besides, first frame pseudo ground truth is utilized to supervise the finetuning process to make accurate inferences.}\label{fig:overview}
\end{figure*}

We evaluate the proposed method on two benchmark datasets, DAVIS 2016 and FBMS-59 datasets. The experimental results demonstrate the state-of-the-art performances on both the two datasets. The main contributions are summarized as follows:
\begin{itemize}
\item First, we introduce a novel hard negative example selection method by incorporating instance proposals, block matching tracklets and motion saliency masks. 
\item Second, we propose a distractor-aware approch to perform the online adaptation to generate video object segmentation with better temporal consistency and avoid the motion error propagation. 
\item Finally, the proposed method achieves the state-of-the-art results on DAVIS 2016 and FBMS-59 datasets with mean intersection-over-union (IoU) scores of 81.6\% and 79.1\%.
\end{itemize}


\section{Related Work}

The related works are introduced from three aspects, semi-supervised VOS, unsupervised VOS, and hard example mining. This section will discuss them accordingly.

\subsection{Semi-supervised video object segmentation.}

Given the manual foreground/background annotations for the first frame in a video clip, semi-supervised VOS methods segment the foreground object along the remaining frames. Deep learning based methods have achieved excellent performance \cite{voigtlaender2017online,cheng2017segflow,jang2017online,yoon2017pixel,xiao2018monet,yang2018efficient}, and static image segmentation \cite{caelles2017one,perazzi2017learning,maninis2017video,huang2018unsupervised,huang2017semantic} is utilized to perform video object segmentation without any temporal information. MaskTrack \cite{perazzi2017learning} considers the output of the previous frame as a guidance in the next frame to refine the mask. OSVOS \cite {caelles2017one} processes each frame independently by finetuning on the first frame, and OSVOS-S \cite{maninis2017video} further transfers instance-level semantic information learned on ImageNet \cite{deng2009imagenet} to produce more accurate results. OnAVOS \cite{voigtlaender2017online} proposes online finetuning with the predicted frames to further optimize the inference network. To fully exploit the motion cues, MoNet \cite{xiao2018monet} introduces a distance transform layer to separate motion-inconstant objects and refine the segmentation results. However, under the circumstances which the object is occluded or changes the movement abruptly, significant performance deterioration will occur. Our approach aims to tackle this challenge using distractor-aware online adaptation.

\subsection{Unsupervised video object segmentation.}

Unsupervised VOS
algorithms \cite{jang2016primary,ma2012maximum,papazoglou2013fast,
wang2015saliency,zhang2013video,li2018instance,li2018unsupervised,hu2018unsupervised,wang2018design} attempt to extract the primary object
segmentation with no manual annotations. Several unsupervised VOS algorithms \cite{grundmann2010efficient,
xu2012streaming} cluster the boundary pixels hierarchically to generate mid-level video segmentations. ARP \cite{koh2017primary} utilizes the recurrent primary object to initialize the segmentation and then refines the initial mask by iteratively augmenting with missing parts or reducing them by excluding noisy parts. Recently, deep learning based methods \cite{tokmakov2017learning,jain2017fusionseg,song2018pyramid,li2018instance,li2018unsupervised} have been proposed to utilize both motion boundaries and saliency maps to identify the primary object. Two-stream FCNs \cite{long2015fully}, LVO \cite{tokmakov2017learninglvo} and FSEG
\cite{jain2017fusionseg}, are proposed to jointly exploit appearance and motion features. FSEG further boosts the performance by utilizing weakly annotated videos, while LVO forwards the concatenated features to bidirectional convolutional GRU. MBN \cite{li2018unsupervised} combines the background from motion-based bilateral network with instance embeddings to boost the performance.


\subsection{Hard example mining.}

There are enormous imbalances between the foreground and background regions since more regions can be sampled from backgrounds than those from foreground. In addition, overwhelming easy negative examples from the background regions have less contributions to train the detector, and thus hard negative example mining approaches are proposed to tackle this imbalanced challenge.

Bootstrapping is exploited in optimizing Support Vector Machines (SVMs) \cite{felzenszwalb2010object} by several rounds of training SVMs to converge on the working set, and modifying the working set by removing easy examples and adding hard examples. Hard example mining is also used in boosted decision trees \cite{dollar2009integral} by training the model with positive examples and a random set of negative examples. The model is then applied to the rest of negative examples to generate false positive examples for retraining the pre-trained model.

Hard negative mining has also been exploited in deep learning models to improve the performance. OHEM \cite{shrivastava2016training} trains region-based object detectors using automatically selected hard examples, and yields significant boosts in detection performance on both PASCAL \cite{everingham2010pascal} and MS COCO \cite{lin2014microsoft} datasets. Focal loss \cite{lin2018focal} is designed to down-weight the loss assigned for well-classified examples and focuses on the training on hard examples. Effective bootstrapping of hard examples is also applied in face detection \cite{wan2016bootstrapping}, pedestrian detection \cite{appel2013quickly}, and tracking \cite{nam2016learning} etc. Both trackers and static image object detectors are applied to select hard examples by finding the inconsistency between the tracklets and object detections from unlabeled videos \cite{tang2012shifting}. In \cite{jin2018unsupervised}, a trained detector is utilized to find the isolated detection, which is marked as a hard negative, from the preceding and following detections. In the proposed approach, we focus on developing an online hard example selection strategy for video object segmentation.


\section{Proposed Method}

The overview of the proposed method is illustrated in Figure \ref{fig:overview}. The proposed method mainly includes three components, 1) both the continuous motion and visual appearance cues are utilized to generate the first frame pseudo ground truth; 2) a novel online hard negative example selection approach is proposed to find high-quality training data; 3) finally, we put forward a distractor-aware online adaptation to facilitate unsupervised video object segmentation.

\subsection{Generate pseudo ground truth} \label{sec:pgt}

Either objectness masks or motion saliency masks cannot generate accurate and reliable initilized mask. Image-based semantic segmentation \cite{chen2018encoder} and instance segmentation \cite{he2017mask} techniques are well developed in recent years. Instead of utilizing semantic segmentation, we apply an instance segmentation algorithm, Mask R-CNN \cite{he2017mask}, to the first frame of the given video without any further finetuning. It is worth pointing out that Mask R-CNN outputs class-specific segmentation masks whereas video object segmentation aims at generating binary class-agnostic segmentation mask. The experiments demonstrate the instance segmentation network produces the segmentation mask with the closest class label to the limited class labels of MS COCO. With further mapping all the classes to one foreground class, the mis-classification has limited influence to the inference process in VOS.

Although instance segmentation provides accurate object regions, the same object may be separated into different parts due to variations of textures, colors, and the effect of occlusions. Besides, only image-based segmentation algorithm is not capable of determining the primary object in a given video clip. Therefore, motion cues are essential to be incorporated to tackle unsupervised video object segmentation. We utilize motion information to select and group instance proposals and then map all the proposals to one foreground mask without knowing the specific category of the object. Specifically, Coarse2Fine \cite{liu2009beyond} is exploited to extract the optical flow between the first frame and second frame of a given video sequence. To avoid the effect of camera motion, we adopt flow-saliency transformation by utilizing a saliency detection method \cite{tokmakov2017learning} on the optical flow to segment moving regions from the background instead of thresholding the flow magnitude. Alternative approaches \cite{baker2011database,zhang2015minimum,xiao2018monet} can be applied to perform the saliency detection. Instance proposals, whose overlapping regions with the motion regions exceed a certain threshold, are selected and grouped as one foreground mask as the pseudo ground truth:

\begin{eqnarray}
P_{GT} = \bigcup_{i = 1}^{n}I_i, \ \ \ (\forall I_i, \  if\ \  \frac{I_i \bigcap M}{I_i} > T)
\end{eqnarray}	

\noindent where $P_{GT}$ represents pseudo ground truth mask, $M$ is the motion mask, $I_i$ represents the $i-th$ instance proposal, $T$ denotes the threshold, and $n$ is the total number of the instance proposals. 

\begin{figure}[t]
\captionsetup[subfigure]{}
\centering 
\subfloat{\includegraphics[width=0.45\linewidth]{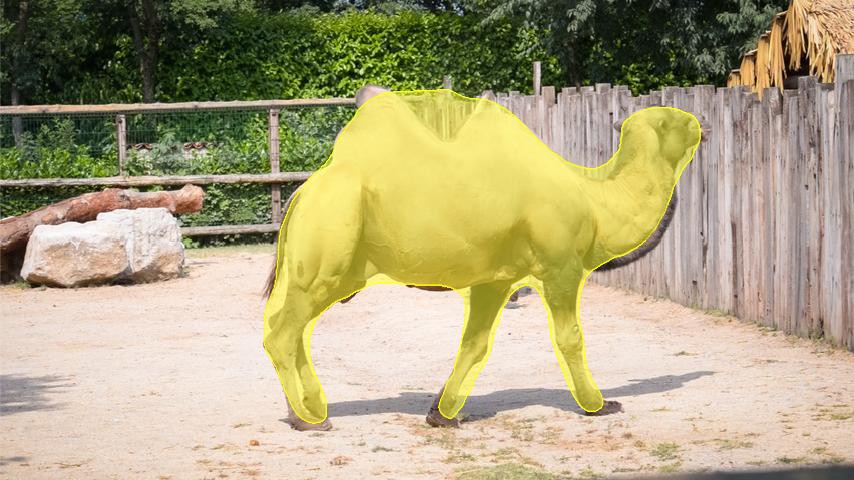}} \hfil 
\subfloat{\includegraphics[width=0.45\linewidth]{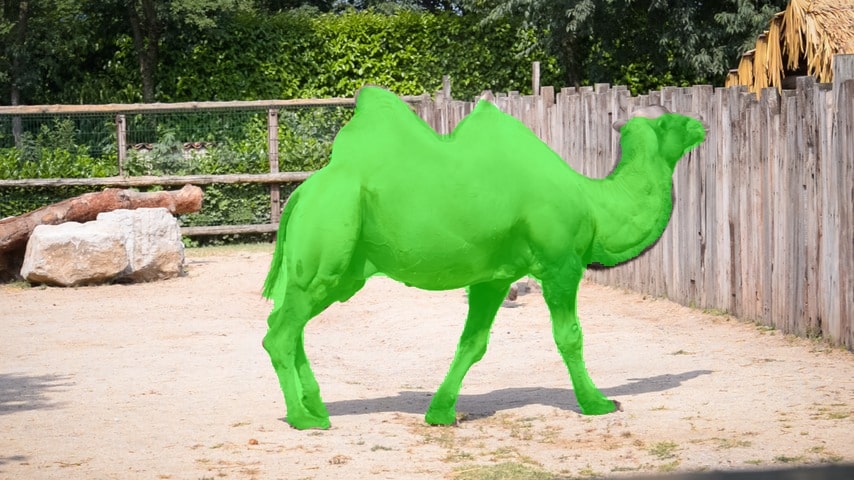}} \hfil \\

\caption{Objectness mask from Mask R-CNN (yellow), Deeplabv3+ (green). Best view in color with 3$\times$ zoom.}\label{fig:compare}
\end{figure}

Semantic segmentation segments the same category of objects with one mask while instance segmentation provides a segmentation mask independently for each instance. It is worth mentioning that the settings of our approach are exploiting instance segmentation instead of semantic segmentation since objects in the same category cannot be distinguished. However, we observed that the pixel-wise labels in the instance segmentation dataset MS COCO are very coarse, while the labels of semantic segmentation datasets, PASCAL and Cityscape \cite{Cordts2016Cityscapes}, are very accurate. Therefore, segmentation networks pretrained on fine-labeled datasets should predict object regions more accurately. To this end, we adopt a semantic segmentation network Deeplabv3+ \cite{chen2018encoder}, pretrained on PASCAL dataset without further finetuning, to generate object masks when the number of each object category is at most one. The number of the objects are determined by the instance segmentation stage. Moreover, as the resolution of the semantic segmentation outputs is lower than that of the VOS datasets, bilinear interpolation and a dense conditional random field (CRF) are utilized to upsample the mask and refine the boundaries, respectively to generate the objectness mask. Sample objectness masks generated from Mask R-CNN and Deeplabv3+ are presented in Figure \ref{fig:compare}, which demonstrates the advantage of exploiting semantic segmentation for object region prediction. Note that more accurate first frame pseudo ground truth leads to higher mean intersection-over-union (mIoU) for video sequences which will be shown in experiments. 

\subsection{Online hard negative/negative/positive example selection} \label{sec:hn}

Although it is straightforward to treat video object segmentation as an image-based foreground/background segmentation without using temporal information, it is difficult to obtain consistent object regions and tackle the truncations, occlusions, and abrupt motion, which leads to inaccurate foreground segmentation. To address this challenge, we propose a novel online process to automatically select hard negative samples using motion cues. Besides, negative and positive examples utilized in the online adaptation are also introduced below.

Hard negatives have similar features with positive examples and the existence of the misalignment between them leads to further mistakes by treating the false positives as positives. The errors are further propagated during the prediction process if the false positives are not effectively suppressed. Therefore, detecting hard negative samples is essential to improve the segmentation accuracy. In Section \ref{sec:pgt}, we fuse the motion mask and the instance proposals to obtain a pseudo ground truth. Similarly, we initialize the hard negatives from the first frame by selecting the detections which are not covered by the motion mask. Note that we only leave the detections with higher or equal confidence score than 0.8 no matter whether the detections are the same category as the positive examples since we observed that when a car is driving across a stop sign, the background stop sign segmentation usually merges into the foreground car segmentation even though they belong to differerent categories. Then, Mask R-CNN is applied to the remaining frames in the same video, and we have both segmentation masks and detection bounding boxes during this process. For each detection in frame $t$, we perform block matching algorithm \cite{dabov2007image} to previous $k$ frames in an enlarged bounding box by $20 \times k$ pixels in each direction, and we empirically select $k$ as 3. For the $k$ previous frames, if the minimum IoU between the instance proposals and matching blocks is at least 0.7, we denote the detections in the current frame as consistent object bounding boxes and the segmentations in the corresponding boxes as consistent object segmentations. Subsequently, if the overlap between the consistent object masks and the motion masks in frame $t$ is below 0.2, we denote these consistent object masks as hard negative examples in frame $t$ which is represented as follows: 

\begin{eqnarray}
HN_S &=& \bigcup_{i = 1}^{n}I_i, \ \ \{\forall I_i, \ \   if \  (I_i \bigcap M < T_1)  \nonumber\\
 & and & \ (\min\limits_{k \in [1, 3]} (f(I_{i_{bb}}) \bigcap P_{i-k})\geq T_2)\}
\end{eqnarray}	 

\noindent where $HN_S$ denotes hard negatives, $M$ is the motion mask, $I_i$ and $I_{i_{bb}}$ represent the instance segmentation and corresponding bounding box respectively, $T_1$ and $T_2$ denote two thresholds, $f$ is the blocking matching function and $P_{i-k}$ is the corresponding bounding box in the $(i - k)^{th}$ frame.

The consistent detections from adjacent frames indicate that these objects are already seen and treated as hard negatives before the current frame, and when the target object moves across the hard negatives, they have larger probability to be distinguished. Our experiments show that significant improvements can be achieved by finetuning the network using these hard negative examples. 

Hard negative examples may be rare in some video sequences, however, it is difficult to get distracted by the easy negative examples in this case. Considering new instances entering the scene are not trained as foreground or background, which may have higher probability to be treated as foreground. We adopt the settings used in \cite{voigtlaender2017online}, each pixel has an Euclidean distance to the closest foreground pixel of the mask. The pixels with a distance larger than the threshold are denoted as negative examples:

\begin{eqnarray}
N_S = \{p_i: \min\limits_{i} E(p_i, Pos) > d \}
\end{eqnarray}	

\noindent where $N_S$ represents the negative examples, $p_i$ represents the pixels in the frame, $d$ denotes the threshold distance, $Pos$ is the positive mask and $E$ is the Euclidean distance. 

\begin{figure}[t]
\captionsetup[subfigure]{}
\centering 
\subfloat{\includegraphics[width=0.8\linewidth]{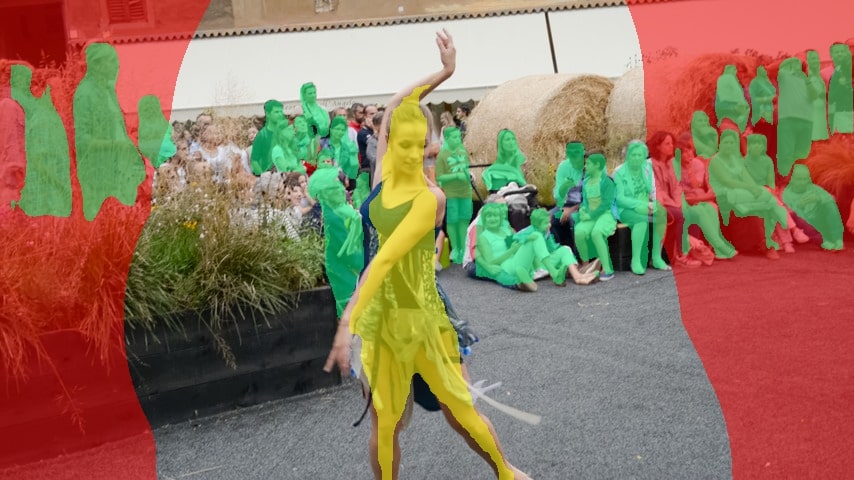}} \hfil 
\caption{Illustration of positive examples (yellow), negative examples (red) and hard negative examples (green).}\label{fig:triple}
\end{figure}

The basic idea of gathering positive examples is to select the high confidence foreground pixels. Considering the motion between consecutive frames, we erode the foreground mask of frame $t - 1$ to initlize the positive examples of frame $t$. However, the objects occluded by the moving object at the beginning have higher probability to be segmented when they are not occluded since the occluded objects are not trained as background and the positive pixels spread to the new objects which are false positives. To tackle this challenge, we regard the intersection of the motion mask and the eroded foreground mask as the positive examples. Note that if there is no intersection between the two masks, which may result in unsatisfactory segmentation, we perform one-shot VOS approach instead of online adaptation for those frames so that the errors from motion will not propagate to the remaining frames. The positive examples with motion cues are as follows:

\begin{eqnarray}
P_t = M \bigcap g(P_{t-1})
\end{eqnarray}	

\noindent where $P_t$ and $P_{t - 1}$ denote the positive masks for the frame $t$ and $t - 1$ respectively and $g$ represents the erosion function. Figure \ref{fig:triple} presents the positive examples, negative examples and hard negative examples.

\subsection{Distractor-aware online adaptation}

In Section \ref{sec:hn}, we propose a process to carefully select positive, negative and hard negative examples for each video sequence. For the purpose of accommodating the foreground object appearance variations across a given video, we perform online adaptation based on the aforementioned positive, negative and hard negative examples and name it as distractor-aware online adaptation (DOA). The model is updated in the current frame and better adapted to make inference in the same frame using DOA. 

Compared with the negative examples, the hard negatives are given higher attention weights, since the positives and hard negatives have similar features and we aim at increasing the discrimination power of distinguishing the objects with similar attributes. The DOA approach can suppress the distractors and provide superior performance compared with that of no adaptation. However, negative examples are also utilized in the DOA approach since the new instance entering into the scene is easily classified as foreground, in addition, sometimes hard negatives do not exist in the frame. Thus, we combine both negative and hard negative examples to finetune the model to suppress the distractor with the loss function: 

\begin{eqnarray}
\mathcal{L}_{curr}= \lambda\mathcal{L}_{hn} + (1-\lambda)\mathcal{L}_{n}
\end{eqnarray}	

\noindent where $\mathcal{L}_{hn}$ and $\mathcal{L}_{n}$ represent pixel-wise segmentation losses for hard negatives and negatives respectively, and $\mathcal{L}_{curr}$ is the loss for the current frame. Here, $\lambda$ is the coefficient for the two losses to balance the contributions of them. When $\lambda = 0$, the loss is equivalent to the loss without hard negatives. As $\lambda$ increases, the loss with easy negatives gets discounted and $\lambda = 0.8$ is set in our experiments when hard negatives exist.

Note that the positive region in frame $t$ is at most the same size of the erosion mask from frame $t - 1$, and thus plays a role as foreground attention, however, more training iterations on the current frame cause inaccurate predictions. We consider incorporating the first frame into the finetuning process since the pseudo ground truth of the first frame can supply high-quality training data. According to the settings in \cite{voigtlaender2017online}, the first frame is sampled more compared with the current frame, and the weight of the loss for the current frame is reduced in order to improve the performance. Subsequently, the joint loss function is as follows:

\begin{eqnarray}
\mathcal{L}_{total}= \alpha\mathcal{L}_{ff} + (1-\alpha)\mathcal{L}_{curr}
\end{eqnarray}	

\noindent where $\mathcal{L}_{ff}$ denotes the loss for the first frame and $\mathcal{L}_{total}$ denotes the total loss. $\alpha$ is the balance coefficient which is set to $0.95$ in our experiments.

The distractor-aware approach completely utilizes the spatial and temporal information from the current frames, the previous several frames and the first frame to set up a long-range consistency over the given video. The finetuning step utilizes high-quality training data to train and thus the inference enables high-quality foreground segmentation results. After the distractor-aware online adaptation, a dense CRF is applied to refine the segmentation.


\section{Experiments}

In this section, the experiments are divided into four parts: datasets and evaluation metrics, implementation details, performance comparison with state-of-the-art and an ablation study. To evaluate the effectiveness of the proposed method, we conduct
experiments on two challenging video object segmentation datasets:
DAVIS 2016 \cite{Perazzi2016} and Freiburg-Berkeley Motion Segmentation (FBMS-59) \cite{ochs2014segmentation}. 

\begin{figure*}[t]
\captionsetup[subfigure]{labelformat=empty}
\centering 
\subfloat{\includegraphics[width=0.187\linewidth]{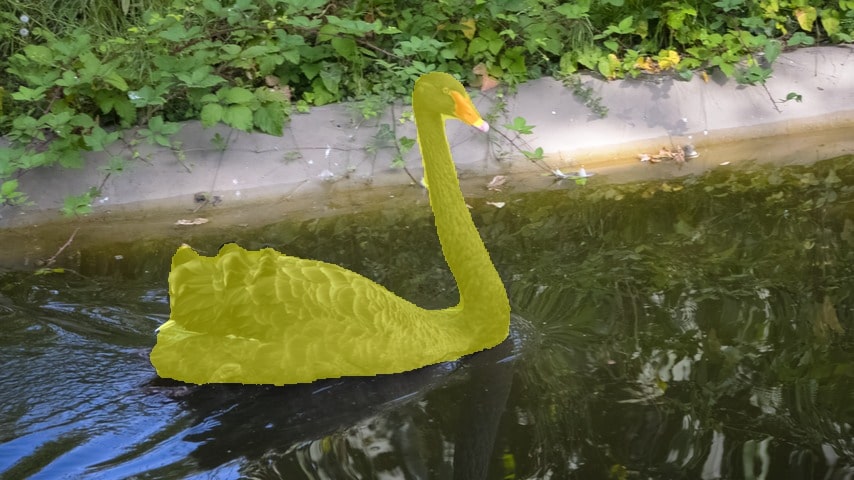}} \hfil
\subfloat{\includegraphics[width=0.187\linewidth]{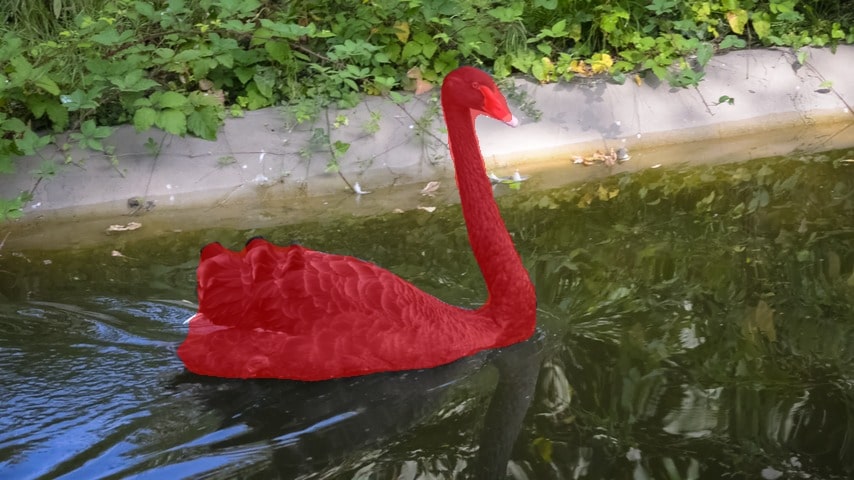}} \hfil
\subfloat{\includegraphics[width=0.187\linewidth]{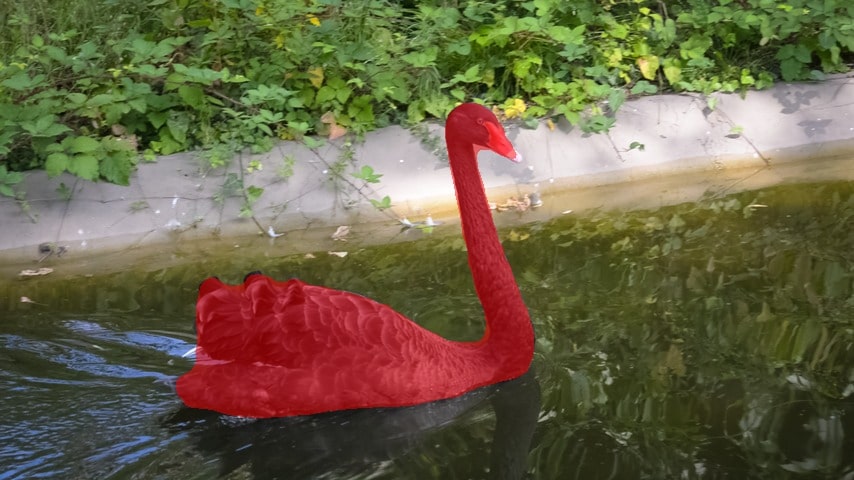}} \hfil
\subfloat{\includegraphics[width=0.187\linewidth]{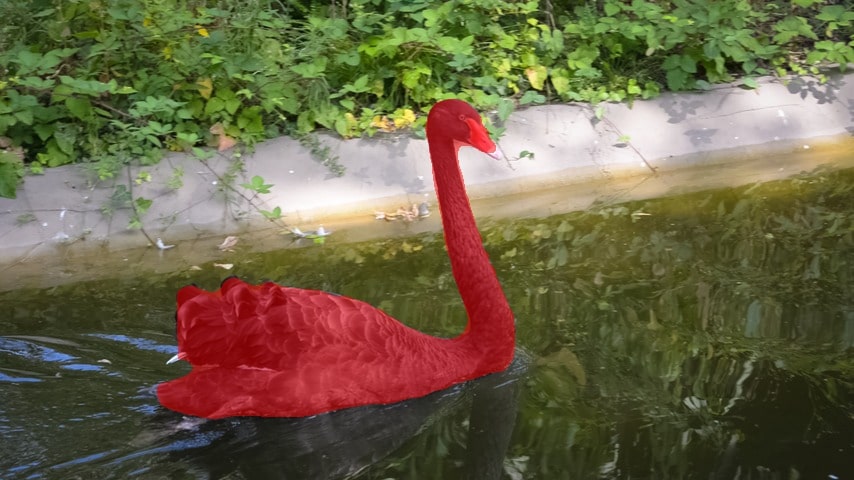}} \hfil
\subfloat{\includegraphics[width=0.187\linewidth]{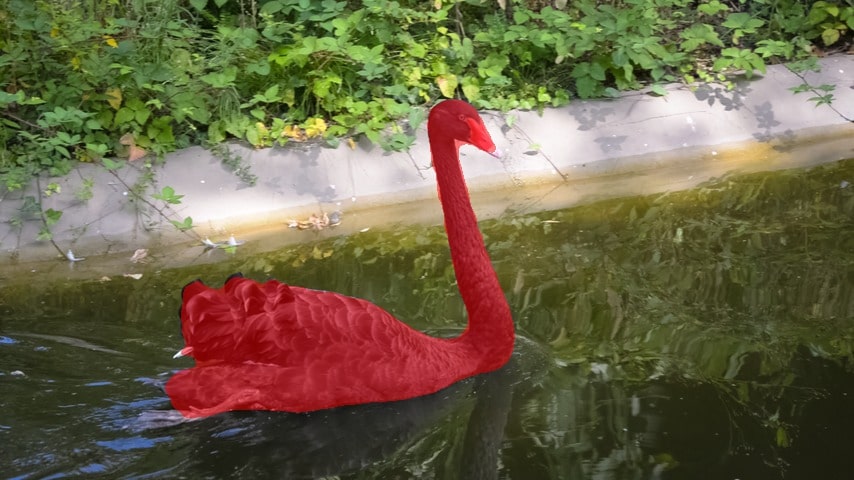}} \hfil \\
\vspace{-0.1in}
\subfloat{\includegraphics[width=0.187\linewidth]{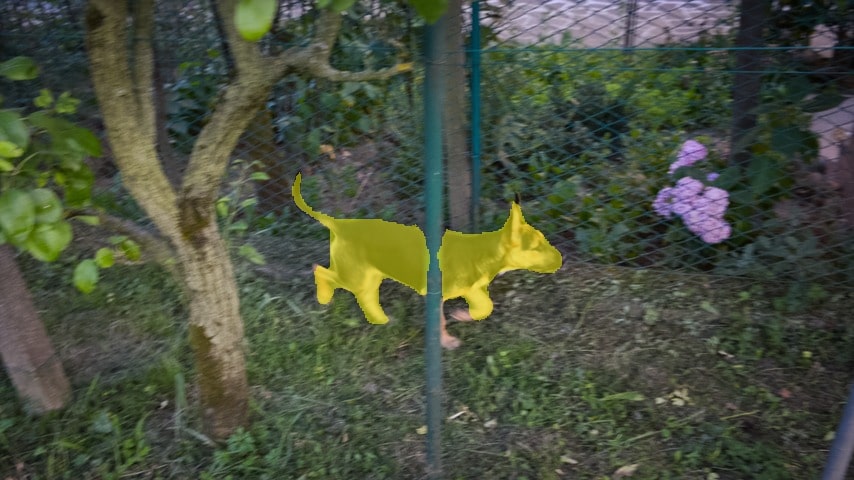}} \hfil
\subfloat{\includegraphics[width=0.187\linewidth]{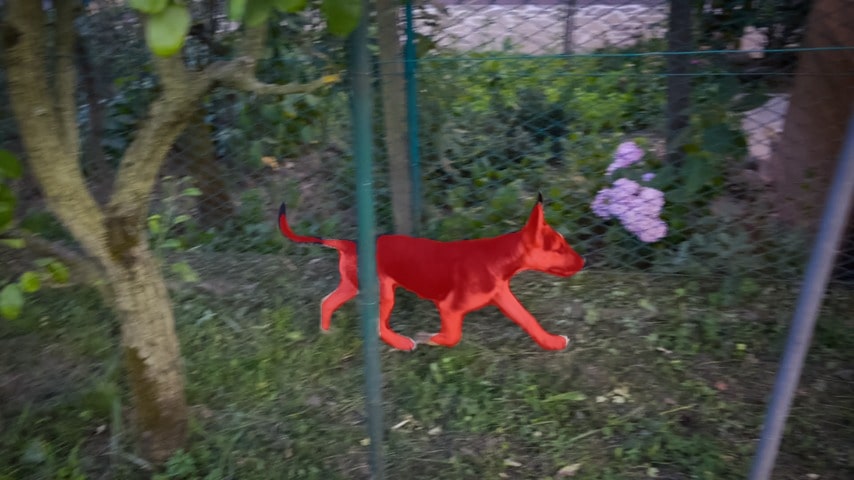}} \hfil
\subfloat{\includegraphics[width=0.187\linewidth]{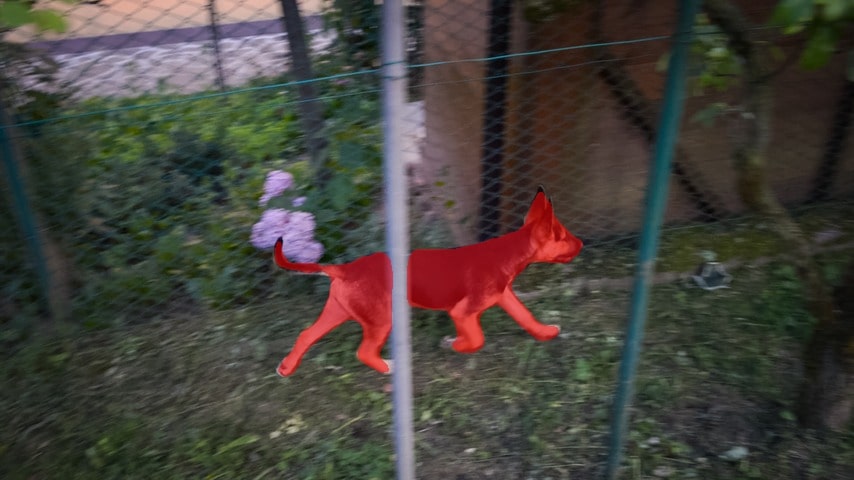}} \hfil
\subfloat{\includegraphics[width=0.187\linewidth]{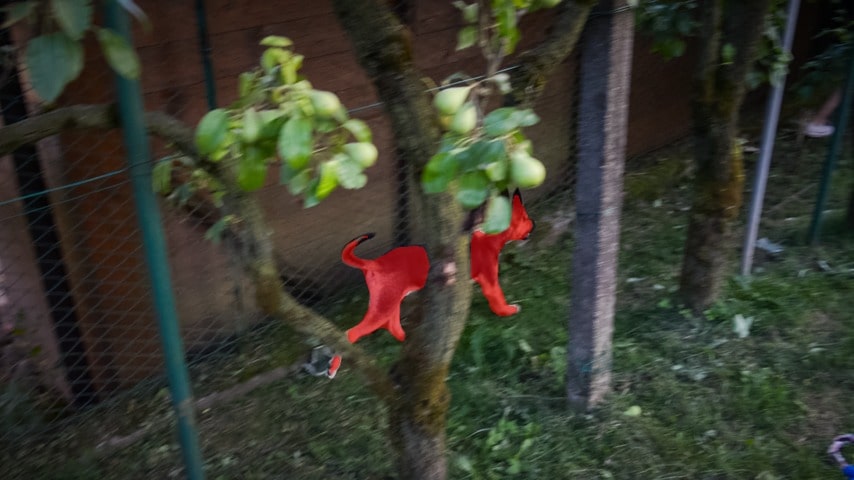}} \hfil
\subfloat{\includegraphics[width=0.187\linewidth]{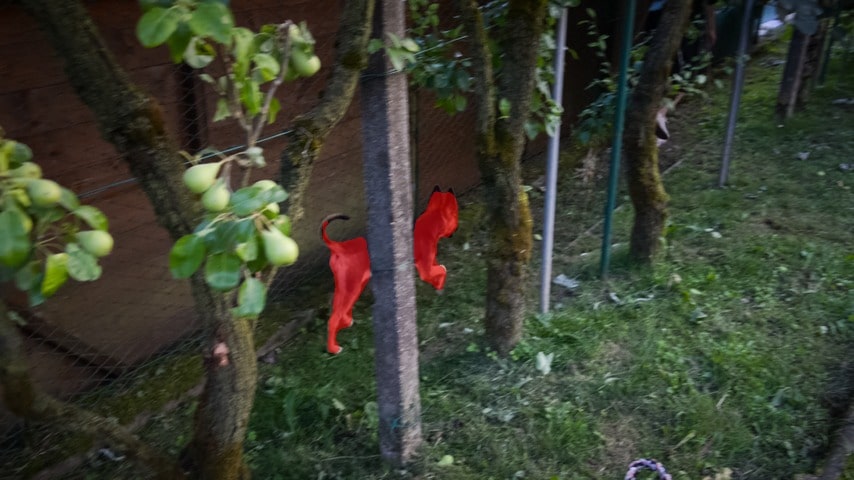}} \hfil \\
\vspace{-0.1in}
\subfloat{\includegraphics[width=0.187\linewidth]{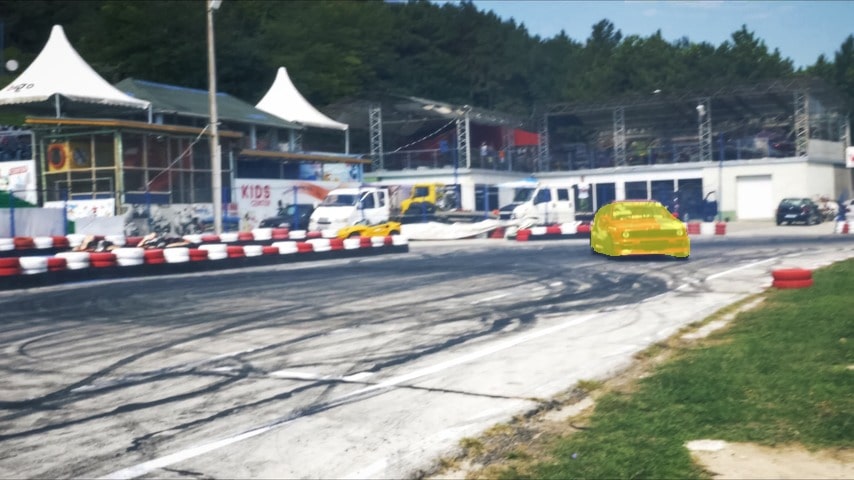}} \hfil
\subfloat{\includegraphics[width=0.187\linewidth]{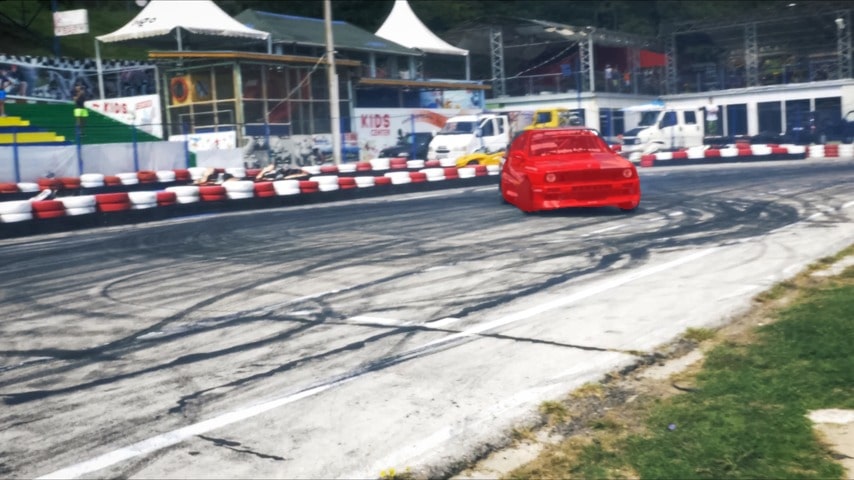}} \hfil
\subfloat{\includegraphics[width=0.187\linewidth]{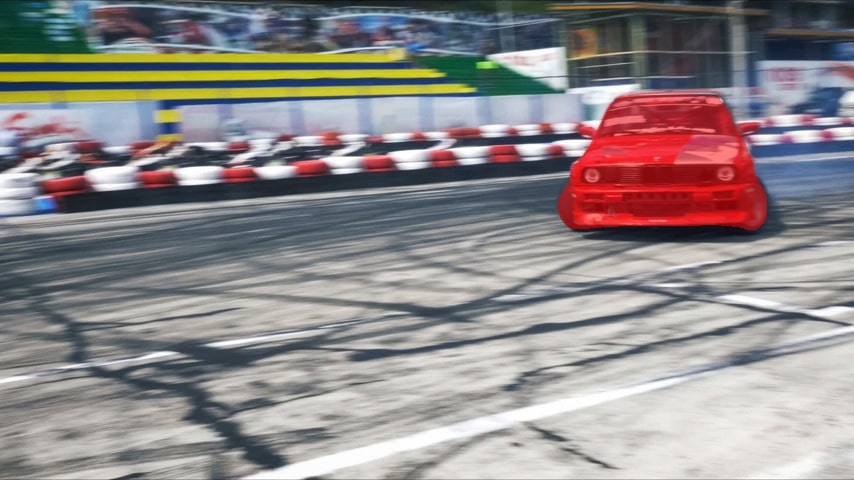}} \hfil
\subfloat{\includegraphics[width=0.187\linewidth]{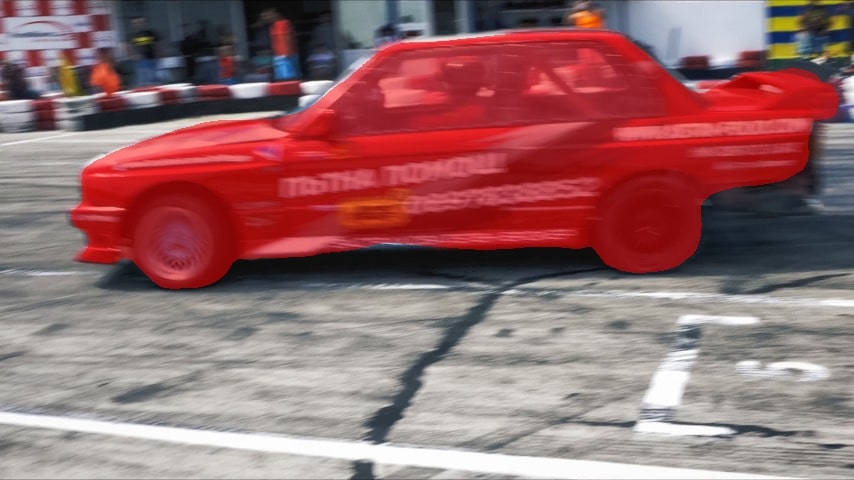}} \hfil
\subfloat{\includegraphics[width=0.187\linewidth]{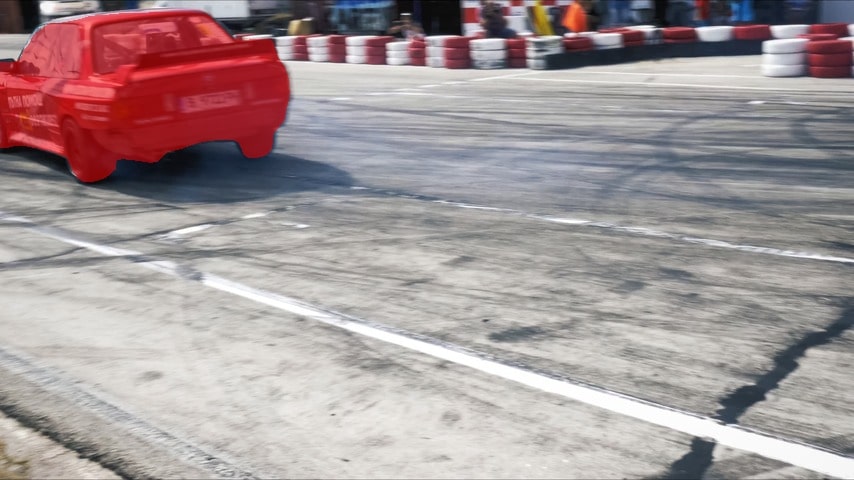}} \hfil \\
\vspace{-0.1in}
\subfloat{\includegraphics[width=0.187\linewidth]{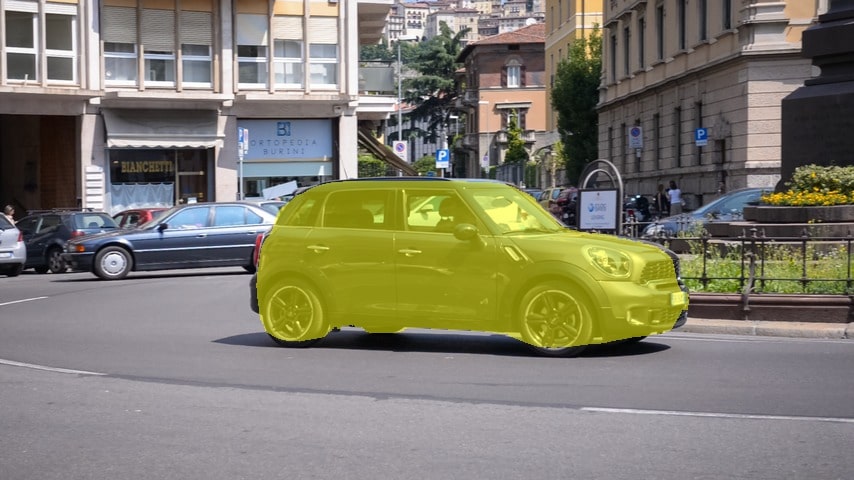}} \hfil
\subfloat{\includegraphics[width=0.187\linewidth]{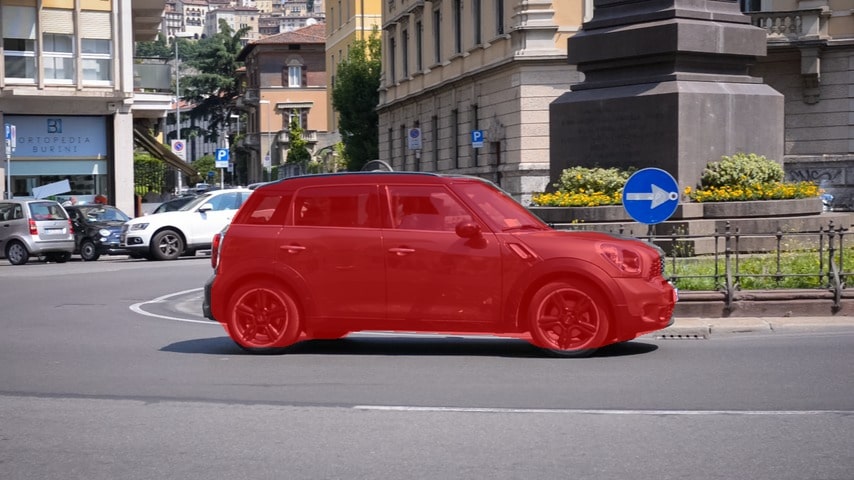}} \hfil
\subfloat{\includegraphics[width=0.187\linewidth]{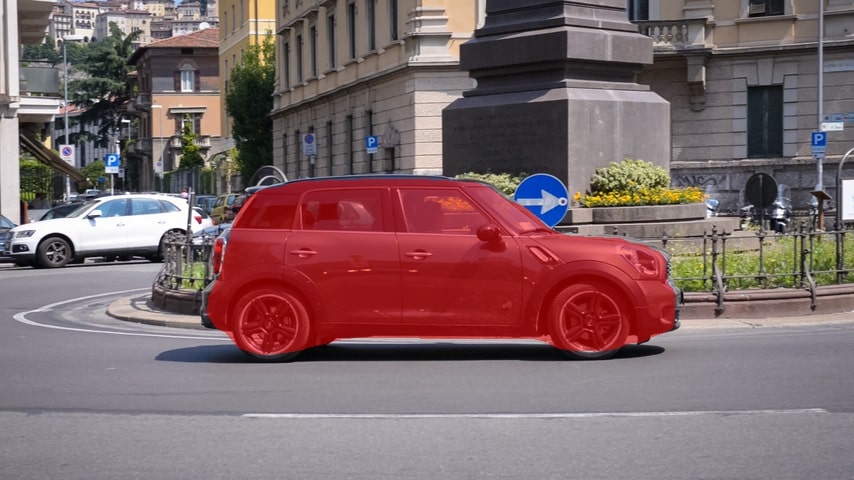}} \hfil
\subfloat{\includegraphics[width=0.187\linewidth]{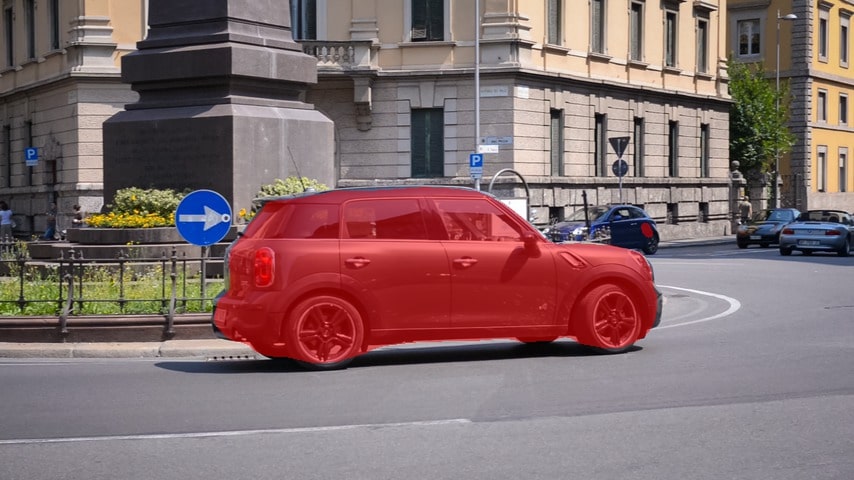}} \hfil
\subfloat{\includegraphics[width=0.187\linewidth]{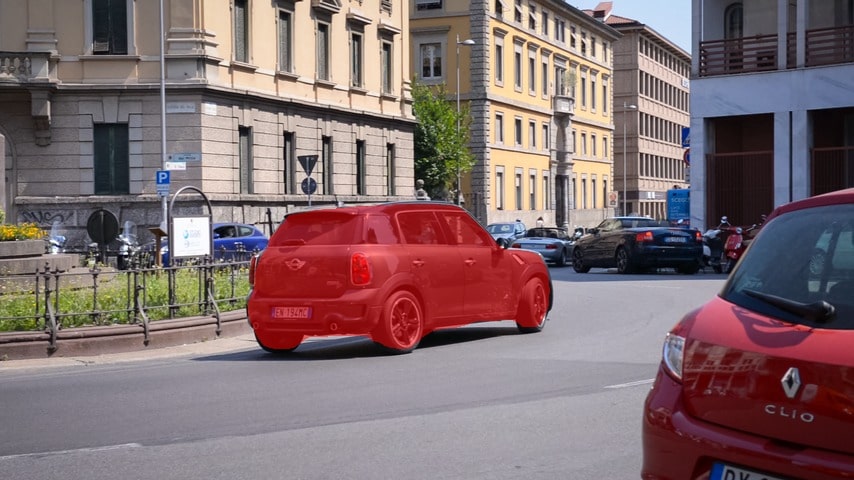}} \hfil \\
\vspace{-0.1in}
\subfloat{\includegraphics[width=0.187\linewidth]{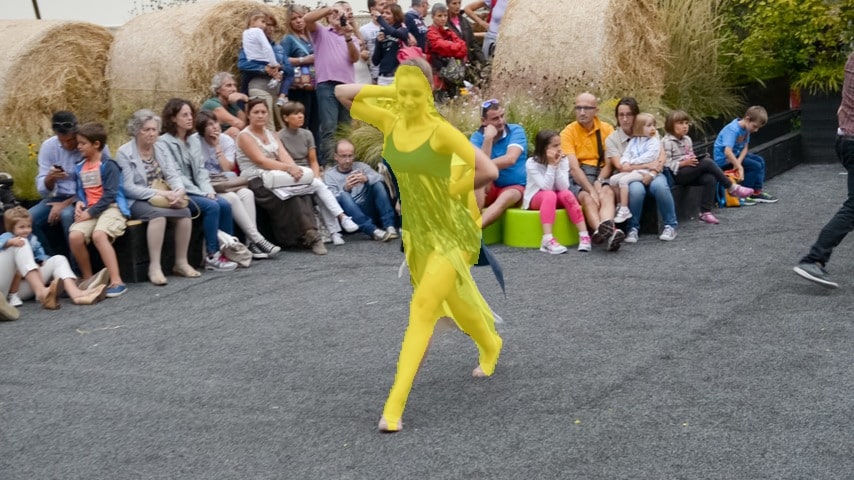}} \hfil 
\subfloat{\includegraphics[width=0.187\linewidth]{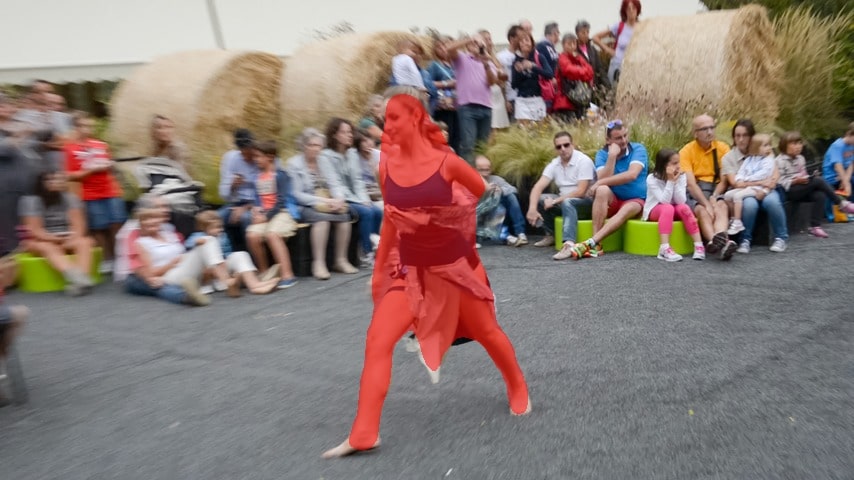}} \hfil 
\subfloat{\includegraphics[width=0.187\linewidth]{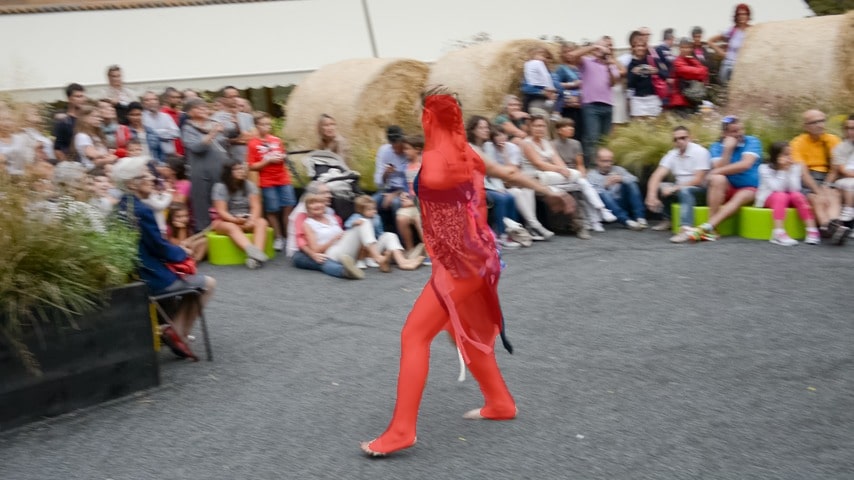}} \hfil
\subfloat{\includegraphics[width=0.187\linewidth]{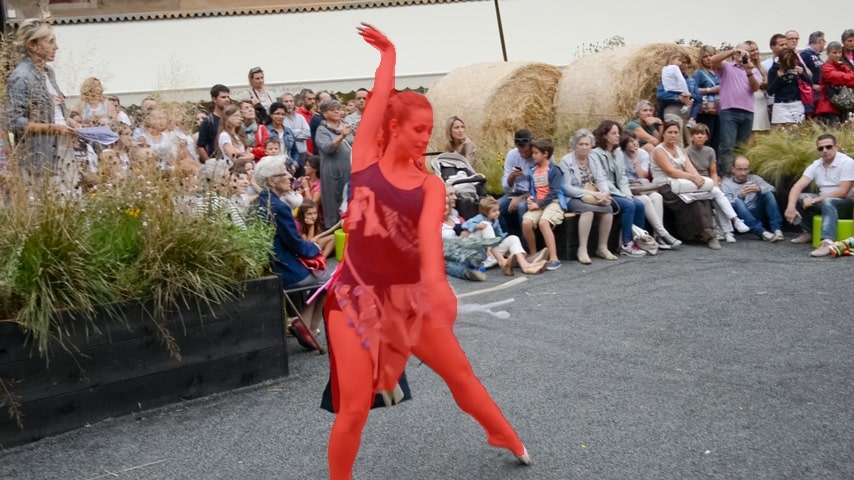}} \hfil
\subfloat{\includegraphics[width=0.187\linewidth]{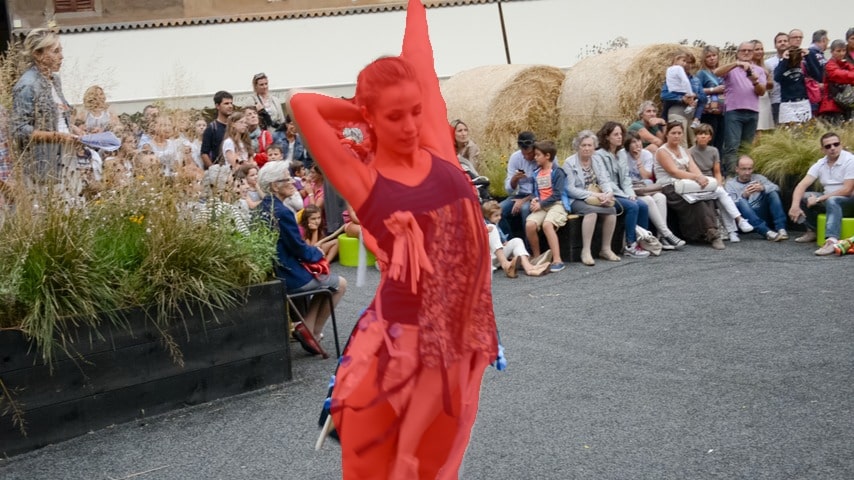}} \hfil\\
\caption{Visual results of the proposed DOA on DAVIS 2016. The pseudo ground truths (in yellow) are illustrated in the first column, and the other columns are the segmentation results (in red) by DOA. The five sequences include the unseen object (first row), strong occlusions (second row), appearance variance (third row), and similar static objects in the background (fourth and fifth row). Best viewed in color with 3$\times$ zoom.}\label{fig:davis}
\end{figure*}

\begin{table*}[t]
\centering
\caption{Comparison of the results of several
methods for the DAVIS 2016 validation dataset. The proposed method outperforms state-of-the-art unsupervised VOS methods and is even better than some supervised VOS approaches in terms of $\mathcal{J}$ Mean and $\mathcal{F}$ Mean ($\%$)} \label{tab:davis}
\begin{tabular}{|c|c|c|c|c|c|c|c|c|c|c|}
\hline
	\multirow{2}{*}{Methods} &
      \multicolumn{3}{c|}{Supervised} &
      \multicolumn{7}{c|}{Unsupervised} \\ \cline{2-11}
 & OnAVOS & OSVOS & MSK & LVO & LMP & FSEG & ARP & IET & MBN & Ours \\ \hline

$\mathcal{J}$ Mean & 86.1 & 79.8 & 79.7 & 75.9 & 70.0 & 70.7 & 76.2 & 78.6 & 80.4 & \bf{81.6} \\ \hline
$\mathcal{F}$ Mean & 84.9 & 80.6 & 75.4 & 72.1 & 65.9 & 65.3 & 70.6 & 76.1 & 78.5 & \bf{79.7} \\ \hline

\end{tabular}
\end{table*}

\subsection{Datasets and evaluation metrics}

We adopt two conventional evaluation metrics, region similarity $\mathcal{J}$ and contour similarity $\mathcal{F}$ to estimate the accuracy.

\noindent
{\bf{Region similarity $\mathcal{J}$.}} Jaccard index $\mathcal{J}$ is defined as intersection-over-union (IoU) between the ground truth mask and the predicted mask to measure the region-based segmentation similarity. Specifically, given a predicted mask $P$ and corresponding ground truth mask $G$, $\mathcal{J}$ is defined as $\mathcal{J} = \frac{P\bigcap G}{P\bigcup G}$.

\noindent
{\bf{Contour similarity $\mathcal{F}$.}} The contour similarity $\mathcal{F}$ is defined as the F-measure between the contour points of the predicted segmentation and the ground truth as proposed in \cite{martin2004learning}. Given contour-based precision $P$ and recall $R$, $\mathcal{F}$ is defined as $\mathcal{F} = \frac{2PR}{P + R}$.

We use DAVIS 2016 and FBMS-59 as two evaluation benchmarks which are introduced below.

\noindent
{\bf{DAVIS.}} The DAVIS dataset is composed of 50 high-definition video
sequences, 30 in the training set and the remaining 20 in the validation set. There are totally 3, 455 densely annotated, pixel-accurate frames. The videos contain challenges such as occlusions, motion blur, and appearance
changes. Only the primary moving objects are annotated in the
ground truth. 

\noindent
{\bf{FBMS.}} The Freiburg-Berkeley motion segmentation dataset is
composed of 59 video sequences with 720 frames annotated. In contrast to
DAVIS, it has multiple moving objects in several videos with instance-level annotations. We do not train on any sequence on FBMS and evaluate using region similarity $\mathcal{J}$ and F-score protocol from \cite{ochs2014segmentation}, respectively. We also convert the instance-level annotations to binary ones by merging all foreground annotations, as in \cite{tokmakov2017learning}.

\subsection{Implementation details}

Motion saliency segmentation and image semantic/instance segmentation are jointly utilized to predict the pseudo ground truth for the first frame. We employ Coarse2Fine \cite{liu2009beyond} optical flow algorithm followed by a flow-saliency transformation approach \cite{tokmakov2017learning} to avoid the effect of camera motion. For instance segmentation, we utilize a re-implementation of Mask RCNN \cite{matterport_maskrcnn_2017} without further finetuning to generate instance proposals in general cases. A semantic segmentation approach Deeplabv3+ \cite{chen2018encoder} is utilized to replace the instance segmentation algorithm Mask R-CNN when the number of objects in each category is at most one. Considering the trade-off between inference speed and accuracy, we utilize the Xception \cite{chollet2017xception} model pretrained on MS COCO and PASCAL with the following settings: output stride is 16, eval scale is 1.0 and no left-right flip. 

We adopt a ResNet \cite{wu2016wider} with 38 hidden layers as the backbone. All implementation and training are based on Tensorflow \cite{abadi2016tensorflow} and ADAM \cite{kingma2014adam} is utlized for optimization. Similar settings with Deeplabv2 \cite{chen2016deeplab} are exploited in this network: large field-of-view replaces the top linear classifier and global pooling layer, and dilations are used to replace the down-sampling operations in certain layers. The ResNet is trained on MS COCO and then finetuned on the augmented PASCAL VOC ground truth from \cite{hariharan2014simultaneous} with a total of 12, 051 training images. Note that all the 20 object classes in PASCAL are mapping to one foreground mask and the background is kept unchanged.

To evaluate on DAVIS 2016 dataset, we further train the network on DAVIS training set and perform two experiments: one-shot finetuning on the first frame with pseudo ground truth, distractor-aware online adaptation with negatives and hard negatives. For both of them, a dense CRF \cite{chen2016deeplab} is applied to refine the semgentation afterwards. For completeness, we also conduct experiments on FBMS-59 dataset, however, we use the PASCAL pretrained model with upsampling instead.

\begin{table*}[t]
\centering
\caption{Comparison of the $\mathcal{J}$ Mean and $\mathcal{F}$ Mean scores ($\%$) of different unsupervised VOS approaches on the FBMS test dataset. Our method achieves
the highest compared with state-of-the-art methods} 
\label{tab:FBMS}
\resizebox{\textwidth}{!}{%
\begin{tabular}{|c|c|c|c|c|c|c|c|c||c|}
\hline
 & NLC \cite{faktor2014video} & FST \cite{papazoglou2013fast} & CVOS \cite{taylor2015causal} & MP-Net-V \cite{tokmakov2017learning} & LVO \cite{tokmakov2017learninglvo} & ARP \cite{koh2017primary} &IET \cite{li2018instance} & MBN \cite{li2018unsupervised}& Ours \\  \hline \hline
$\mathcal{J}$ Mean & 44.5 & 55.5 & - & - & - & 59.8 & 71.9& 73.9 & \bf{79.1} \\ \hline
F-score & - & 69.2 & 74.9 & 77.5 & 77.8 & - & 82.8 & 83.2 & \bf{85.8} \\ \hline
\end{tabular}
}
\end{table*}

\begin{table}[t]
\centering
\caption{Ablation study of the three modules in distractor-aware online adaptation: (1) negative example addition (+N), (2)
hard negative example addition (+HN), and (3) fusion of positive mask with motion mask (+MP), assessed on the DAVIS 2016 validation set} \label{tab:ablation}
\begin{tabular}{|cccc|l|}
\hline
 +N & +HN & +MP & CRF & $\mathcal{J}$ Mean \\ \hline
- & - & - & - & 76.7  \\ \hline
\checkmark & - & - & - & 78.9 {\scriptsize{\color{red} +2.2}} \\ \hline
\checkmark & \checkmark & - & - & 80.1 {\scriptsize{\color{red} +1.2}} \\ \hline
\checkmark & \checkmark & \checkmark & - & 80.6 {\scriptsize{\color{red} +0.5}}\\ \hline
\checkmark & \checkmark & \checkmark & \checkmark & 81.6 {\scriptsize{\color{red} +1.0}} \\ \hline
\end{tabular}
\end{table}

\subsection{Performance comparison with state-of-the-art}

\begin{table}[t]
\centering
\caption{Comparison of the first frame influence for the DAVIS 2016 validation dataset. We finetune on the first frame and perform inference for the remaining frames without online adaptation. We compare the performances ($\%$) from the pseudo ground truth generated from Mask R-CNN ($PGT_M$) and jointly from Mask R-CNN and Deeplabv3+ ($PGT_{MD}$), the erosion and dilation masks from $PGT_M$, and ground truth mask (GT)} \label{tab:ff}
\resizebox{\columnwidth}{!}{%
\begin{tabular}{|c|c|c|c|c|c|}
\hline
 & Erosion & Dilation & $PGT_M$ & $PGT_{MD}$ & GT \\ \hline
First frame $\mathcal{J}$ Mean & 67.9 & 74.9 & 79.0 & 81.3 & 100  \\ \hline
The whole val set $\mathcal{J}$ Mean & 65.0 & 73.8 & 75.8 & 76.7 & 80.4  \\ \hline
\end{tabular}
}
\end{table}

\noindent
{\bf{DAVIS 2016.}} The performances on DAVIS 2016 are summarized in Table \ref{tab:davis}. The proposed DOA approach outperforms state-of-the-art unsupervised VOS techniques, e.g., LVO
\cite{tokmakov2017learninglvo}, and ARP \cite{koh2017primary}, FSEG \cite{jain2017fusionseg}, LMP \cite{tokmakov2017learning}, IET \cite{li2018instance} and MBN \cite{li2018unsupervised}. Specifically, the superior gaps to the second-best MBN are 1.2\% and 1.2\% in terms of $\mathcal{J}$ Mean and $\mathcal{F}$ Mean. Moreover, the proposed method provides convincing performance when compared to some recent semi-supervised VOS techniques such as OSVOS \cite{caelles2017one} and MSK \cite{perazzi2017learning}. The qualitative results of the proposed DOA are presented in Figure \ref{fig:davis}. The first column is the first frame with pseudo ground truth annotation. Our approach yields encouraging results in challenging sequences. The blackswan in the first row is an unseen object category in the training data, our approach is shown to cope well and generate accurate segmentation masks. The second row shows that our algorithm works well for strong occlusions and the third row shows that our method produces accurate segmentation masks for considerable non-rigid deformations. The bottom two rows show our approach produces robust predictions when there are multiple distractors and messy background by exploiting distractor-aware online adaptation. 

We also evaluate the influence of the pseudo ground truth by utilizing the pseudo ground truth generated from Mask R-CNN ($PGT_M$) and jointly from Mask R-CNN and Deeplabv3+ ($PGT_{MD}$). The first frame is finetuned and inferred for the remaining frames without online adaptation. The ground truth and the erosion and dilation masks of ($PGT_M$) are also compared in Table \ref{tab:ff}. We observe that the performance for a video clip is highly correlated with the overlap ratio in the first frame of this video.

\noindent
{\bf{FBMS-59.}} The proposed method is evaluated on the FBMS-59 test set with 30 sequences in total. The results are presented in Table \ref{tab:FBMS}. Our method outperforms the second best method in both evaluation metrics, with $\mathcal{J}$ Mean of 79.1\% and $\mathcal{F}$ Mean of 85.8\% which are 5.2 \% and 2.6\% higher than the second best MBN \cite{li2018unsupervised}, respectively.

\subsection{Ablation studies}

We study the four major components of the proposed methods and then summarize the effects of the components including negative examples, hard negative examples, fusion of motion mask and positive mask, CRF, in Table \ref{tab:ablation}. The baseline without online adaptation is a ResNet trained on the PASCAL dataset and DAVIS 2016 training set. The negative examples provide 2.2\% enhancement over the baseline in terms of $\mathcal{J}$ Mean. The hard negative examples combined with negative examples further improve the performance by 1.2\%, which demonstrates our online adaptation approach is effective when dealing with confusing distractors and hard negatives and negatives are complementary. Subsequently, the fusion of motion mask and positive mask brings further 0.5\% boost since motion information helps select the positive mask to avoid the effects of distractors occluded by the positive mask at the beginning. Finally, additional CRF post-processing is combined to boost the performance by 1.0\%.

\begin{figure}[t]
\captionsetup[subfigure]{}
\centering 
\subfloat{\includegraphics[width=0.48\linewidth]{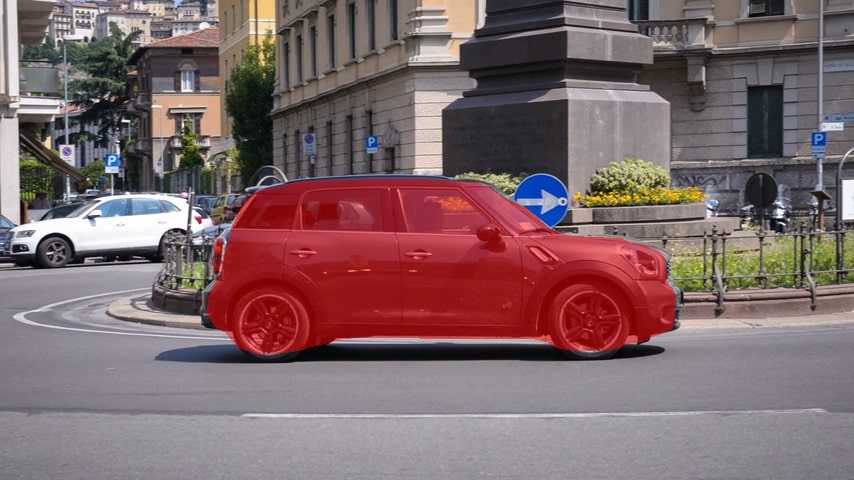}} \hfil 
\subfloat{\includegraphics[width=0.48\linewidth]{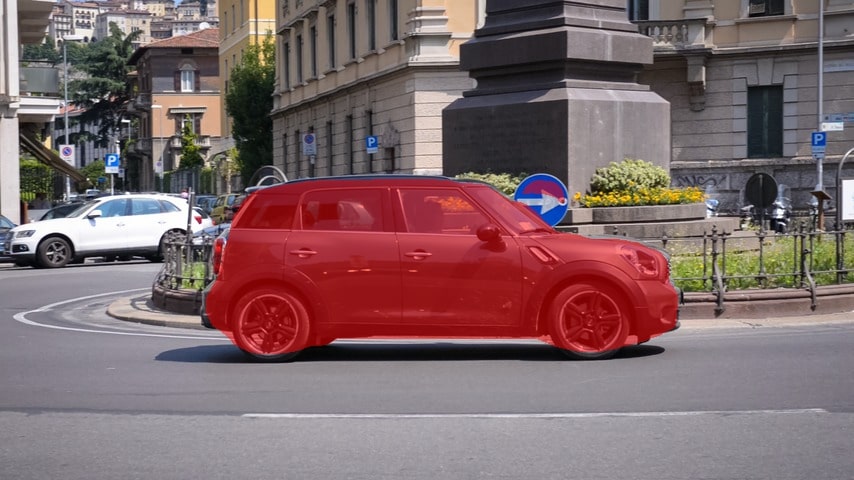}} \hfil \\
\vspace{-0.12in}
\subfloat{\includegraphics[width=0.48\linewidth]{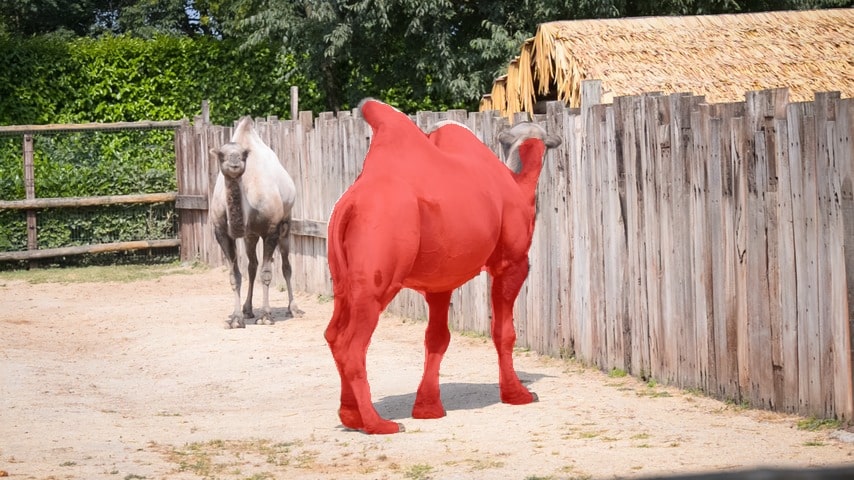}} \hfil 
\subfloat{\includegraphics[width=0.48\linewidth]{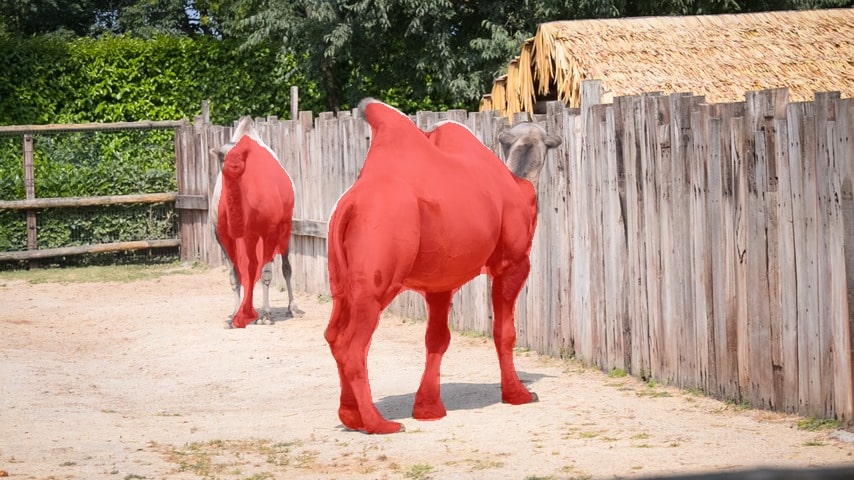}} \hfil 

\caption{Comparison of qualitative results on the key components of online adaptation. The first row presents the differences of w/ HN (left) and w/o HN (right), and the second row presents the differences of w/ MP (left) and w/o MP (right). Best viewed in color.}\label{fig:online}
\end{figure}

Figure \ref{fig:online} shows the qualitative performances with different components. We compare the effect of hard negative examples for the ``car-roundabout'' sequence in the first row, and the positive examples influences for the ``camel'' sequence in the second row. The segmentation with hard negatives training ignores the stop sign while that without hard negatives finetuning merges the stop sign segmentation into the car, which indicates the importance of hard negatives finetuning. To investigate effectiveness of the fusion of motion mask and positive mask eroded from the previous frame (MP), we compare the qualitative results with MP and without MP on the ``camel'' sequence where the walking camel is easily distinguished from the static camel with MP.

\section{Conclusion}

A distractor-aware online adaptation for unsupervised video object segmentation is proposed. The motion between adjacent frames and image segmentation are combined to generate the approximate annotation, pseudo ground truth, to replace the ground truth of the first frame. Motion-based hard example mining and block matching algorithm are integrated to produce distractors which are further incorporated into online adaptation. In addition, motion-based positive examples selection is combined with the hard negatives during online updating. Besides, we conduct an ablation study to demonstrate the effectiveness of each component in our proposed approach. Experimental results show that the proposed method achieves state-of-the-art performance on two benchmark datasets.


{\small
\bibliographystyle{ieee}
\bibliography{egbib}
}

\end{document}